\documentclass{article}

\PassOptionsToPackage{numbers}{natbib}
\usepackage[final]{neurips_2021}

\usepackage[utf8]{inputenc} % allow utf-8 input
\usepackage[T1]{fontenc}    % use 8-bit T1 fonts
\usepackage{hyperref}       % hyperlinks
\usepackage{url}            % simple URL typesetting
\usepackage{booktabs}       % professional-quality tables
\usepackage{amsfonts}       % blackboard math symbols
\usepackage{nicefrac}       % compact symbols for 1/2, etc.
\usepackage{microtype}      % microtypography
\usepackage{xcolor}         % colors
\usepackage{amsmath,mathtools}

\usepackage{graphicx}

\title{A good body is all you need: avoiding catastrophic interference via agent architecture search}

\author{
  Joshua Powers \\
  \texttt{jpowers4@uvm.edu}
  \AND
  Ryan Grindle \\
  \texttt{rgrindle@uvm.edu}
  \And
  Lapo Frati \\
  \texttt{lfrati@uvm.edu}
  \And
  Josh Bongard \\
  \texttt{jbongard@uvm.edu}
}

\begin{document}

\maketitle

\begin{abstract}
In robotics, catastrophic interference continues to restrain policy training across environments. Efforts to combat catastrophic interference to date focus on novel neural architectures or training methods, with a recent emphasis on policies with good initial settings that facilitate training in new environments. However, none of these methods to date have taken into account how the physical architecture of the robot can obstruct or facilitate catastrophic interference, just as the choice of neural architecture can. In previous work we have shown how aspects of a robot’s physical structure  (specifically, sensor placement) can facilitate policy learning by increasing the fraction of optimal policies for a given physical structure. Here we show for the first time that this proxy measure of catastrophic interference correlates with sample efficiency across several search methods, proving that favorable loss landscapes can be induced by the correct choice of physical structure. We show that such structures can be found via co-optimization---optimization of a robot's structure and control policy simultaneously---yielding catastrophic interference resistant robot structures and policies, and that this is more efficient than control policy optimization alone. Finally, we show that such structures exhibit sensor homeostasis across environments and introduce this as the mechanism by which certain robots overcome catastrophic interference.
\end{abstract}
\section{Introduction}
Catastrophic interference is a phenomenon that occurs when training policies on multiple environments simultaneously: a training step that results in improvement in one environment causes a greater reduction in performance in other environments. This comes with various consequences, one of which are policies that specialize to only one of the environments. This problem is often related to the problem of catastrophic forgetting \cite{french1999catastrophic,goodfellow2013empirical,kirkpatrick2017overcoming,masse2018alleviating}, however in catastrophic forgetting environments are trained sequentially, where in catastrophic interference they are trained simultaneously as stated.

Many of the methods used to combat catastrophic intereference rely solely on advances to neural architecture or training methods \cite{zhang2020differentiable}. Most are focused on making changes to small subsets of neural network weights \cite{kirkpatrick2017overcoming,masse2018alleviating,french1991using,robins1995catastrophic,he2018overcoming,beaulieu2018combating,schwarz2018progress,titsias2019functional}. Others have attempted to sidestep the problem by learning good initial weights such that they can be quickly updated when switching between tasks \cite{finn2017model,gidaris2018dynamic}.

None of these methods however consider the effects that the architecture of the robot housing the policy has on this phenomenon. The robot's body is often considered a part of the environment \citep{ha2019reinforcement}. From this perspective, it is the only part of the task environment that is changeable prior to or during training. Consequently, it has been shown in various task settings that an appropriate robot design can simplify the problem of learning a sufficient control policy \cite{lichtensteiger1999evolving,vaughan2004evolution,brown2010universal,bongard2011morphological,kriegman2019automated,PERVAN2019197}. However, so far these findings have been restricted to examples with a single training environment; they did not need to overcome multi-domain problems such as catastrophic interference. In a multitask setting, \citet{powers2018effects} recently demonstrated that certain body plans suffer catastrophic interference, while others do not. It was hypothesized there that a robot with an appropriate design could in some cases alias separate tasks: certain designs are able to move so that seemingly different environments converge sensorially to a common and familiar observation. This is not unlike tilting one's head to recognize a familiar face in a rotated image. However, this conjecture was not isolated and tested. Likewise, the relationship between the body and the loss landscape was not investigated.

Here, we provide a more thorough investigation of the role of embodiment in catastrophic interference, based on the assumption that in order to avoid catastrophic interference, there must exist a set of policy parameters that yield adequate performance across several task environments. Since a robot's mechanical design can change which sets of policy parameters are appropriate for each individual task environment, we here test the hypothesis that a specific physical property of the robot's design---sensor distribution---can help or hinder learning by inducing more or less overlap in specialized policy parameter sets across multiple task environments. These regions of overlap thus correspond to sets of general purpose policy parameters.

We first define metrics to explicitly measure the distribution and number of specialized and general policy parameter sets for a given robot design. We then tested whether those metrics actually correlate with sample efficiency for different training algorithms, and found that they do for all of the tested algorithms. We show how robot designs that facilitate learning can be found by co-optimizing policy and sensor distribution parameters. We found that co-optimization results in significantly better sample efficiency than policy optimization alone. Lastly, we show that the designs found during co-optimization follow a sensor homeostasis gradient. This leads to the hypotheses that part of what allows an agent's body to facilitate policy search is its ability to facilitate sensor homeostasis, and that including homeostasis in future loss functions may further improve sample efficiency in multi domain policy and agent design training.

\section{Methods}
The code implementation an related materials for the following section can be found at \url{https://github.com/jpp46/ARXIV2021.git}.

\subsection{Experimental Overview}
\label{sec:overview}
A minimal agent, task, and environment were employed to test the hypothesis that co-optimizing policy and structure parameters increases sample efficiency during training: agents are trained to approach a light source, and they are placed in environments with differing light source locations.

\begin{figure}
    \centering
    \includegraphics[width=0.3\linewidth]{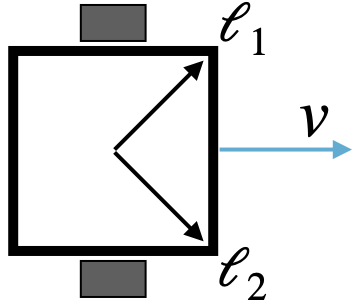}
    \caption{The robot is modeled as having a square frame with two separately-driven wheels and two infrared sensors located on the dorsal surface of the frame at locations defined by two vectors emanating from the center of the frame $\ell_1$ and $\ell_2$. The vector $v$ shows the forward direction. This design is a hand designed baseline based on a Braitenberg Vehicle \cite{braitenberg1986vehicles}.}
    \label{fig:robot}
\end{figure}

The robot is modeled as having a square frame (edge length $0.5m$) with two separately-driven wheels and two infrared sensors. A baseline, hand designed robot can be seen in Fig.~\ref{fig:robot}.
The sensors detect light according to the inverse square law, i.e., $1/d^2$, where $d$ is the distance from the light source and are modeled without occlusion. The motors driving the wheels are contralaterally connected to the sensors by weighted synapses yielding two trainable parameters $w_1, w_2 \in [-1.0, 1.0]$.

In this paper we explore change to a single design attribute: the physical location of the two sensors, which can be placed anywhere on the dorsal surface of the robot's square body.
The location of the $i$-th sensor $\ell_i$ can be described by its Cartesian coordinates $\ell_i = (x, y)$, where $x, y \in [-0.5, 0.5]$, and (0, 0) denotes the center of the body (Fig.~\ref{fig:robot}).

If we define $\theta$ as the space of all possible synapse weight pairs ($w_1$, $w_2$), we can investigate how a given sensor distribution ($\ell_1$, $\ell_2$) affects $\theta$.
Since we cannot perform an exhaustive sweep over the infinitude of possible sensor positions, we discretize each dimension of $\ell_i$ into nine uniformly-spaced bins. Because sensors are varied in two dimensions ($x$ and $y$) there are $9^2=81$ possible locations for each sensor. Since there are two sensors, and each can be in 81 possible discrete locations, the total design space consists of $81^2=6561$ possible robot designs.

For each of these 6561 designs, we conducted another sweep over the synapse weights $(w_1, w_2)$, likewise discretizing each weight into 121 evenly-spaced values within $[-1,1]$, yielding $121^2 = 14641$ different weight controllers per robot design. For each of the $6561\times14641=96059601$ robot design and policy pairs, we evaluated the simulated robot for a phototaxis task across four environments as described below in \S\ref{sec:simulation}. Doing a grid search as described creates an exhaustive investigation of the weight space ($\theta$) for each robot design, thus allowing an analysis as described in \S\ref{sec:metrics} to measure the predicted ability of a robot design to facilitate policy training. We then  applied four different training algorithms to each robot design (see \S\ref{sec:training}) to test for any correlation between the utility of a given design predicted by our metrics and actual performance from training policies on it. Finally, we compare policy optimization using the default design against simultaneous co-optimization of design and policy (see \S\ref{sec:cooptimization}), and analyze the resulting optimized designs.

All of the experiments were run in parallel on a cpu computing cluster that is powdered by $95\%$ renewable energy. In total the experiments took $564,786$ cpu hours.

\subsection{Simulating the robot.}
\label{sec:simulation}
The robot is trained on a phototaxis task across four environments: the robot should move toward a light source in each environment. In each environment a light is placed in the center of an $(x,y)$ plane and the robot is placed diagonally at a distance of $8$ body lengths away from the light source; specifically at positions $(d, d)$, $(d, -d)$, $(-d, d)$, $(-d, -d)$ where $d=\frac{4}{\sqrt{2}}$ constitutes the four differing environments. Success in an environment is determined by how close the robot comes to the light source. We assume the light source has a radius ($r=0.075m$) that determines its physical size, and maximum loss is defined to be reached if the robot touches the light source during the course of its evaluation.

Given the robot's and environment's simplicity, rather than using a simulated robot and task environment, the location and orientation of the robot in the plane can be defined by a system of differential equations, where the change in position and orientation is determined by the change in light captured by two sensors. Ignoring deviations from the idealized environment, such as sensor noise and friction, the rate of angular and linear velocities will be proportional to a linear combination of the sensor values.

Let $\alpha(t)$ denote the angle of the robot at time $t$, where $\alpha=0$ denotes the positive $x$ direction, and $\phi(t) = (x(t), y(t))$ denote the position of the robot in the world. If the robot is located at the origin and facing east ($\alpha = 0$), its two light sensors are located exactly at $\ell_1$ and $\ell_2$, and they each capture some amount of light $s_1(t)$ and $s_2(t)$, respectively. 

Hence the absolute position of the $i$-th sensor is $\phi(t) + R_\alpha \ell_i^{\hspace{3pt}T}$, where
	\begin{equation}
		R_\alpha = \begin{bmatrix}
			\cos \alpha & -\sin \alpha \\
			\sin \alpha & \cos \alpha \\
		\end{bmatrix}
	\end{equation}
is the two-dimensional counterclockwise rotation matrix (in the amount $\alpha$). 

Since we formulate the problem such that it is the robot's initial position and heading that is adjusted in each environment instead of the position of the light source, we can assume that the light source is always at the origin. Then, the distance of $\ell_i$ from the light source is given by:  \mbox{$\|\phi(t) + R_\alpha \ell_i^{\hspace{3pt} T}\|$.}
And since the intensity of light is inversely proportional to the square of the distance, the sensor values are given by: 
\begin{equation}
\label{eq:sensor_value}
s_i(t) = \|\phi(t)^T + R_\alpha {\ell_i}^T\|^{-2},
\end{equation}

The robot turns based on the difference between the sensor values multiplied by the two neural network weights, the velocity of the robot is equal to the average of the two sensor values. 
Thus, the following determines the location and orientation of the robot:
	\begin{equation}
	    \label{eq:ode}
		\begin{cases}
			\dot{x} = v(t) \cos \alpha	\\
			\dot{y} = v(t) \sin \alpha \\
			\dot{\alpha} = w_1s_1(t) - w_2s_2(t), \\
		\end{cases}
	\end{equation}
where $v$ is the velocity of the robot given by $2v(t) = w_1s_1(t)+w_2s_2(t).$

\subsection{Metrics}
\label{sec:metrics}
We here define two metrics $M_L$ and $M_{CI}$ that are measured over the $K=4$ environments. These metrics measure how a robot design impacts the weight space of the control policy and consequently provide a prediction of sample efficiency, if policy training were performed using that design. $M_L$ measures controller learnability: how easy it would be to learn a generalist controller. $M_{CI}$ measures resistance to catastrophic interference: the probability that starting training with an environment-specific controller will result in discovery of a generalist controller.

\begin{figure}
    \centering
    \includegraphics[width=0.3\linewidth]{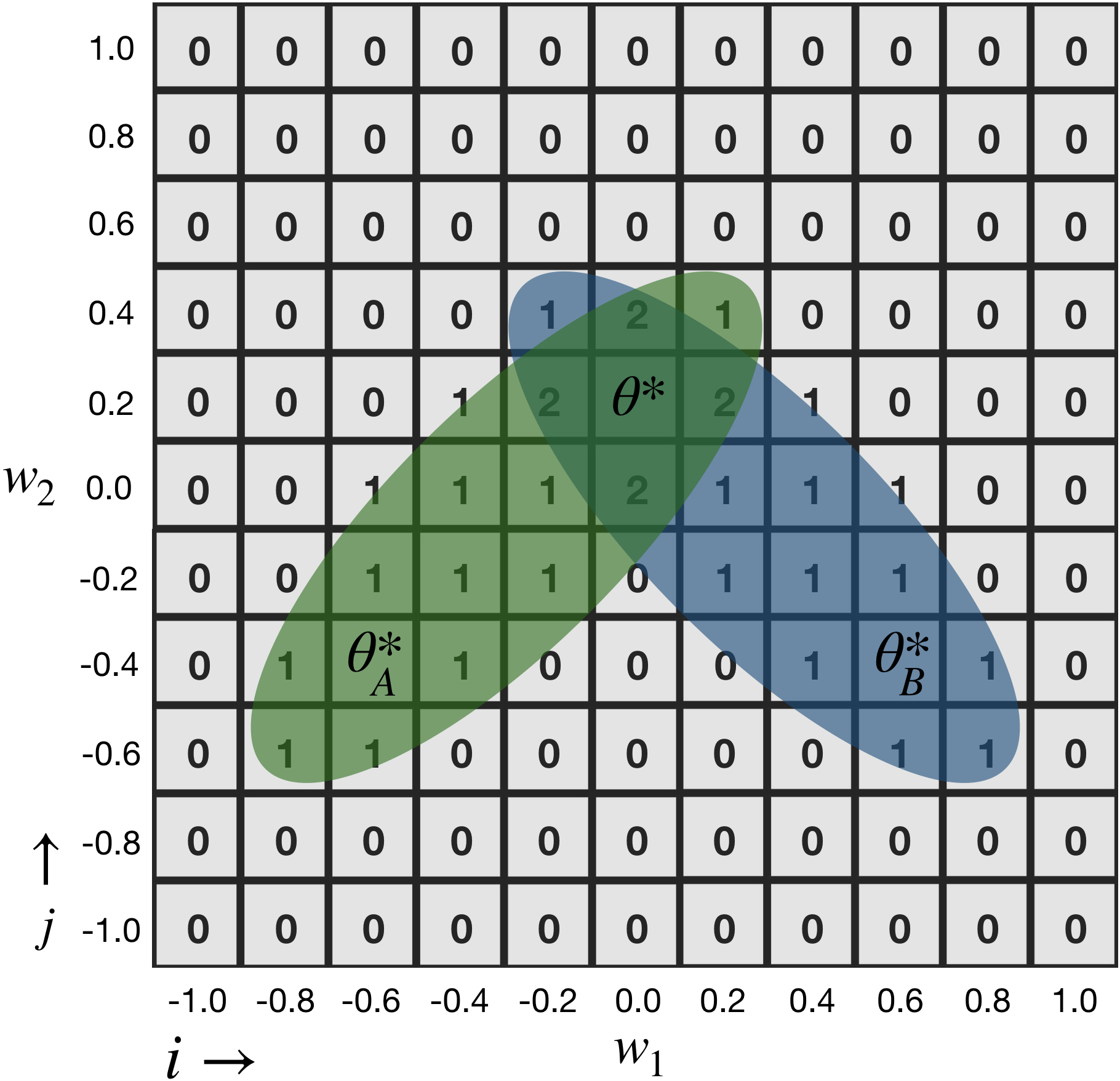}
    \caption{An example of overlapped binary success matrices for two environments A and B. Each element represents a different set of policy parameters. Generalist controllers lie within the intersection $\theta^*$ of successful environment-specific controllers $\theta_k^*$.}
    \label{fig:metrics}
\end{figure}

Given a design $(\ell_1, \ell_2)$ and environment $k$, a binary success matrix $S^{(k)}(\ell_1, \ell_2)$ is constructed such that each element $S^{(k)}_{i, j}(\ell_1, \ell_2)$ is either 1 (the light source was touched) or 0 (it was not) and each element corresponds to particular values of weights $w_1$ and $w_2$. By overlapping the success matrices for a fixed design across the four environments, we can visualize the manifolds $\theta_k^*$ where $k\in\{1, 2, 3, 4\}$ for the robot (Fig.~\ref{fig:metrics}).

We define the overlap $\mathcal{O}$ as an element-wise sum of the success matrices over each environment $k$:
\begin{equation}
    \label{eq:overlap}
    \mathcal{O} = \sum_{k=1}^K S^{(k)}(\ell_1, \ell_2) .
\end{equation}

The learnability metric is simply the proportion of $|K|=4$ values, which denote success in all environments, in the overlapped success matrices, compared to the entire matrix space:
\begin{equation}
    \label{eq:metric1}
    M_L = \frac{g_{|K|}(\mathcal{O})}{n^2},
\end{equation}
where $g_k$ is a function that counts the total elements of a matrix with value equal to $k$ and $n$ is the square dimension of the matrix defined by the discrete policy parameter set sweep.

Resistance to catastrophic interference is measured by:
\begin{equation}
    \label{eq:metric2}
    M_{CI} = \begin{cases}
    0 \; \text{ if } \mathcal{O} \text{ is a null matrix}, \\
    g_{|K|}(\mathcal{O}) \left[\sum_{k=1}^K g_k(\mathcal{O})\right]^{-1}
    \; \text{otherwise}.
    \end{cases}
\end{equation}
which is the number of policy parameter sets that solved all $K$ environments divided by the number of sets that solved at least one.

\subsection{Training}
\label{sec:training}
We used four different training methods to tune the weights of the control policies: random search, generating set search, separable natural evolution strategies (NES), and differential evolution (DE). These methods were chosen because they encompass different paradigms in optimization with the first (random search) serving as a baseline. Generating set search is considered a powerful derivative free algorithm from the direct search paradigm \cite{kolda2003optimization}. Natural evolution strategies is another derivative free method loosely based on biological evolution \cite{wierstra2014natural}. Differential evolution is also based on biological evolution but is generally considered to be in a different class of evolutionary based optimizers than NES \cite{price2013differential}. Thus, together, these methods give motivation that the results from this research extend beyond the actual methods used\footnote{All of the search methods were used as is from \url{https://github.com/robertfeldt/BlackBoxOptim.jl} using the default hyper parameters from the code base.}.

We applied each training method to each of the $6561$ robot designs, five times using different initial random policies. For each method the loss function was computed as the sum of the minimum distance from the light source achieved in every environment. During training we also kept track of the total number of environments in which the robot was successful. We then compute the average number of evaluations it took to achieve success in all environments for a given training method and plot them against the designs' metrics $M_L$ and $M_{CI}$ to measure the Pearson correlation and corresponding p-value. A strong negative correlation is desirable here: designs with higher metrics require a lower number of evaluations to achieve optimality. This would indicate that good designs predicted by the metrics yielded significantly better sample efficiency.

\subsection{Co-Optimization}
\label{sec:cooptimization}
To simultaneously optimize both the controller and sensor placement of the robot we use the Borg Multiobjective Evolutionary Algorithm (Borg MOEA) \cite{hadka2013borg}. As before the goal of the algorithm is to produce a policy that works in all four environments in as few evaluations as possible. However rather than optimize directly on the sum of score in each environment we treat each of these scores as it's own objective, and pick the best solution at the end based on the aggregate sum of the objectives. The algorithm works on 6 trainable parameters: the 2 values for the first light sensor vector, the 2 values for the other light sensor vector, and the 2 weight values of the controller.

We compare this to the same algorithm, but fix the design of the robot to the baseline design in Fig.~\ref{fig:robot}. In this baseline the algorithm only optimizes 2 trainable parameters which are the 2 weights of the controller.

We run each algorithm over 30 different random seeds and take the average number of evaluations to required to achieve success in all four environments. We then run a Mann Whitney U Test to compare if there were any significant differences between the two methods and report a corresponding p-value.

Lastly, we analyze the sensor values experienced by designs found over the course of co-optimization with dynamic time warping (DTW) \cite{giorgino2009computing}. The dynamic time warping algorithm is a state-of-the-art way to measure the similarity between two signals by determining the cost to align the signals via stretching and/or shifting. For each environment we record a vector of the light sensor signals experienced (four in total). For one of the light sensor we compute the average DTW distance between all combinations of experienced signals (six comparisons) to get a signal score for the light sensor, we then do the same for the other light sensor. We average both of those scores to get the aggregate DTW distance for the design, the lower this number is the more similar the light sensor signals were between environments.
\begin{figure*}[h]
    \centering
    \includegraphics[scale=0.5]{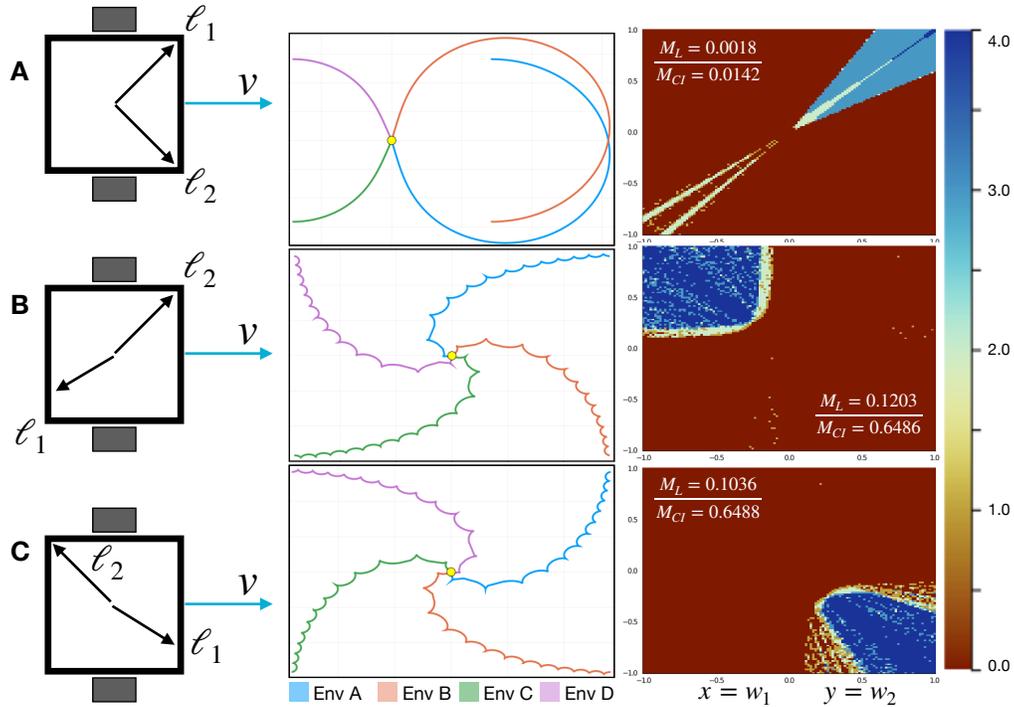}
    \vspace{-40pt}
    \caption{Examples of different robot designs with corresponding policy trace, success matrices, and metrics. The vectors $\ell_1$ and $\ell_2$ show the placement of the light sensors for a specific design. The colors in the second column show the trace of the $(x, y)$ coordinates of the robot in the four different starting positions as it moves toward the light source represented by the yellow point. The last column shows the weight assignments that solve different environments; red assignments solve no environments, while any other color solves at least one, and blue assignments solve all four environments. \textbf{(A)} The baseline morphology as shown in Fig.~\ref{fig:robot}. \textbf{(B)} The design with the best learnability metric ($M_L$). \textbf{(C)} The design with the best resistance to catastrophic forgetting ($M_{CI}$)}
    \label{fig:metrics_results}
\end{figure*}

\begin{figure*}[h]
    \centering
    \includegraphics[width=0.5\linewidth]{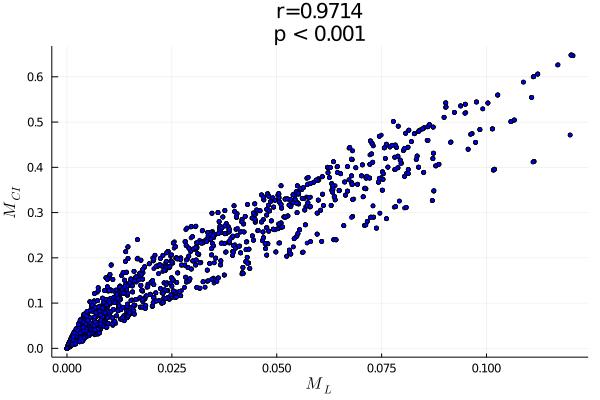}
    \caption{The correlation between design learnability and resistance to catastrophic interference. Each point is a design with its learnability on the x-axis and resistance on the y-axis. The two variables are strongly correlated in this case. The density of points is more concentrated near the origin showing that high scoring designs are relatively rare.}
    \label{fig:ml_mci_correlation}
\end{figure*}

\begin{figure*}[!t]
    \centering
    \textbf{Random Search}\\
    \includegraphics[width=0.45\linewidth]{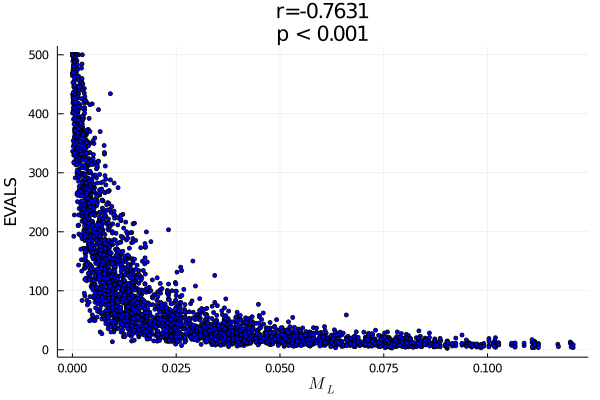}
    \includegraphics[width=0.45\linewidth]{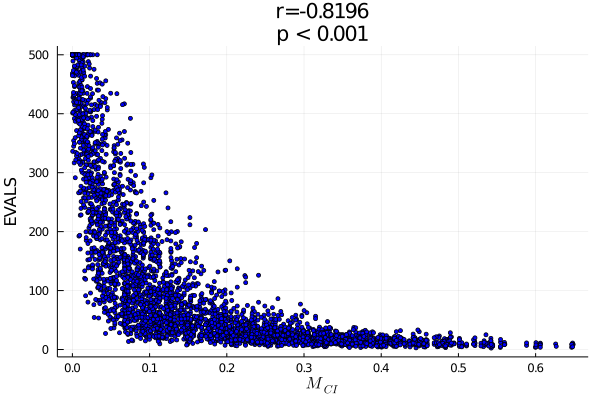}
    
    \textbf{Generating Set Search}\\
    \includegraphics[width=0.45\linewidth]{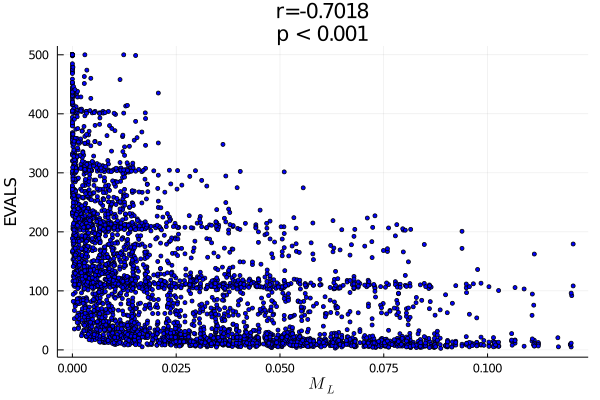}
    \includegraphics[width=0.45\linewidth]{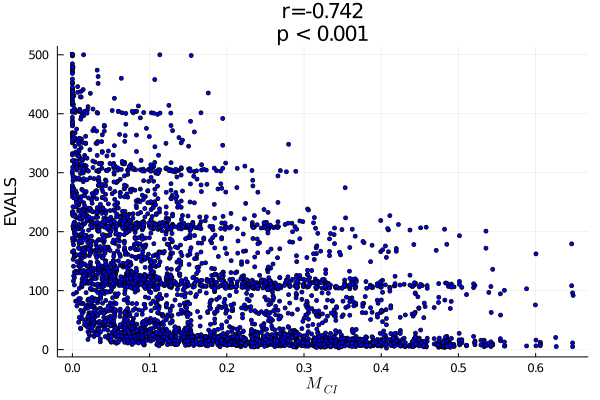}
    
    \textbf{Natural Evolution Strategy}\\
    \includegraphics[width=0.45\linewidth]{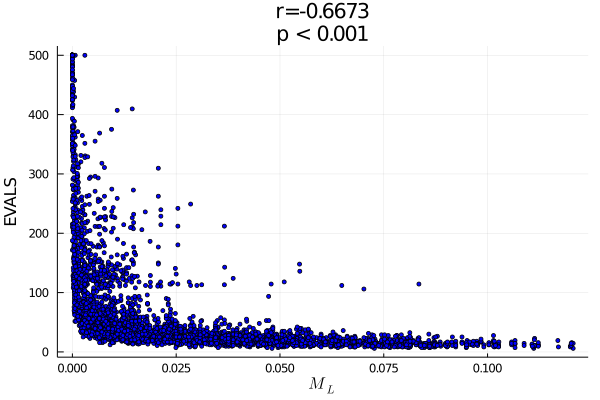}
    \includegraphics[width=0.45\linewidth]{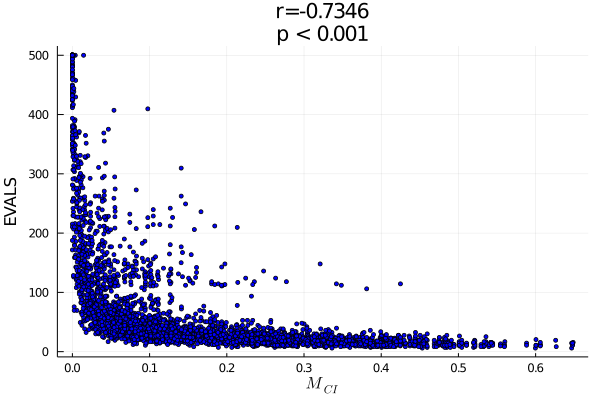}
    
    \textbf{Differential Evolution}\\
    \includegraphics[width=0.45\linewidth]{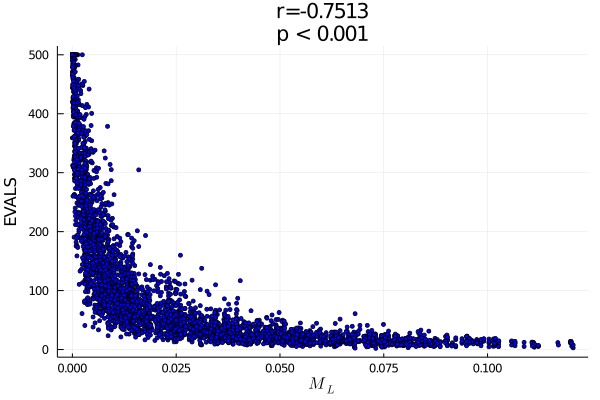}
    \includegraphics[width=0.45\linewidth]{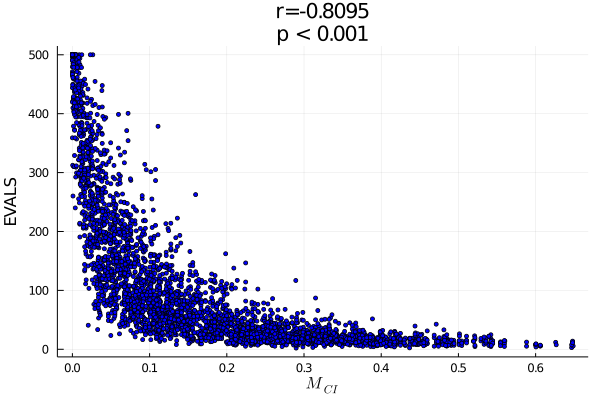}
    \caption{For each design we plot sample efficiency against the design's learnability ($M_L$) and resistance to interference ($M_{CI}$) scores across the search methods. In each case there is a significant negative correlation demonstrating that a design's score is predictive of the required effort to train a policy for that design.}
    \label{fig:evolution_correlation}
\end{figure*}

\begin{figure*}[h]
    \centering
    \includegraphics[width=0.45\linewidth]{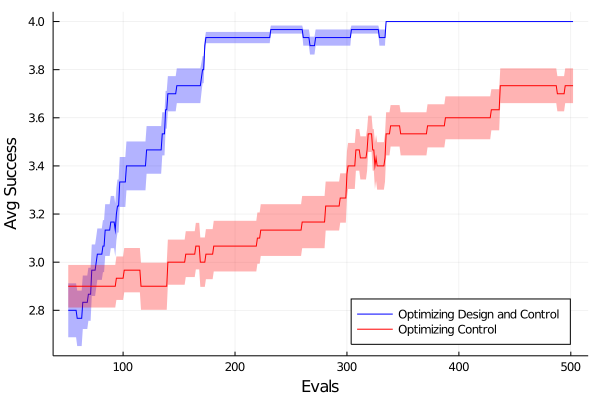}
    \includegraphics[width=0.45\linewidth]{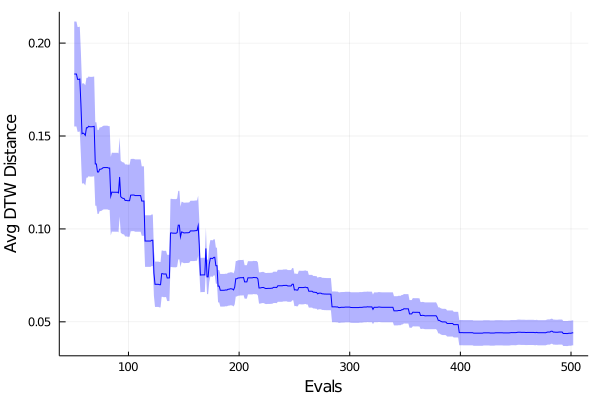}
    \caption{\textbf{(Left)} Average number of environments a policy is successful in (Average Success) as a function of evaluation time, for policy optimization using the baseline design (red) and co-optimization of design and control (blue). Co-optimization was significantly better that controller optimization alone. \textbf{(Right)} The average Dynamic Time Warping (DTW) distance between the robots' sensors in all of the environments for the robot designs found during co-optimization. As optimization progresses, designs progress from those that experience environments differently (high DTW) to those that experience differing environments more similarly (low DTW). Shading in both panels reports the $95\%$ confidence interval.}
    \label{fig:cooptimization}
\end{figure*}

\section{Results}
We employed the DifferentialEquations.jl package \cite{christopher_rackauckas_2021_4785349} for numerical integration of the robot's location and orientation using standard Euler Integration with a time step of $0.1$ for $10^5$ time steps.

For each evaluated mechanical design and policy (sensor locations and synapse weights), the robot's trajectory is computed in each of the four environments defined in \S\ref{sec:simulation}. Again, if robot's trajectory comes within 0.075 meters of the light source, the robot is determined to have succeeded in that environment. Otherwise, it is determined to have failed.

The mechanical design sketched in Fig.~\ref{fig:metrics_results}B had the highest learnability score, with $M_L = 0.1206$. A very similar design (Fig.~\ref{fig:metrics_results}C) scored the best in resistant to catastrophic interference: the proportion of resistant to nonresistant policies for that design was $M_{CI}=0.6479$. In general the best policies for both metrics featured a specific similar design feature of asymmetric placement of the light sensors. The baseline design (Fig.~\ref{fig:metrics_results}A) had a much lower learnability ($M_L=0.0018$) and resistance to catastrophic interference ($M_{CF}=0.0142$) then the best found asymmetrical designs.

The worst designs for both metrics have scores of zero and feature designs where the lights sensors are both placed in the same location or extremely close to one another. This makes particular sense because the closer the light sensors are the less differential there is between signals thus making it difficult or impossible (in the case of overlapping placement) for the robot to distinguish between different environments with out some sore of behavioral memory mechanism such, as a recurrent network.

In Fig.~\ref{fig:metrics_results} the successes matrices of weight manifolds for all the mentioned designs can be seen in detail, where dark blue represents weight assignments that succeed in all for environments for a given design. These weight manifolds show clearly that in this case the optimal weight assignments for the asymmetrical designs should be much easier to find during training, while the baseline design is akin to looking for a needle in a haystack. This also shows that for this specific robot it appears a high score in one metric correlates with the others. We specifically tested this by plotting the metric scores for all of the designs along with the correlation between the two metrics. As shown in Fig.~\ref{fig:ml_mci_correlation} there is a significant positive correlation ($p < 0.001$) between the metrics, thus giving an explanation for why the best designs under both metrics are so similar.

We can also see that based on the frequency of metrics $M_L$ and $M_{CF}$ that there are many designs (including intuitive symmetric ones) that score poorly on $M_L$ and $M_{CF}$ while there are relatively few designs that perform well.

When we conducted the training of the policies for each of these designs as described in \S\ref{sec:training}, we found that a design's score over the metrics was significantly correlated with the number of evaluations required to solve the phototaxis task for all four environments. This can be seen in Fig.~\ref{fig:evolution_correlation}. On the x-axis the metric is plotted and the y-axis is the average number of evaluations required to achieve success in all four environments, thus we see a strong negative correlation across the methods with a p-value less that $0.001$, meaning that a design's metric score was predictive of the required effort to train that design on the task.

Lastly, we compared co-optimization of the policy and robot design as described in \S\ref{sec:cooptimization} versus policy optimization of the baseline design. As can be seen in Fig.~\ref{fig:cooptimization}, co-optimization significantly outperformed ($p < 0.001)$ policy optimization alone, despite increased complexity in the dimension of optimizable parameters. In fact, in the number of evaluations attempted, optimizing the policy for the baseline design never achieved competence in all four environments, and thus, that design ultimately succumbed catastrophic interference. We further analyzed the designs found by co-optimization (Fig.~\ref{fig:cooptimization}) and discovered that they progressively move from designs with high DTW distance scores to ones with low DTW scores. Thus without any guidance, the optimization naturally found a homeostatic design (one that sensorally experiences differing environments similarly) or a design which was best suited to optimization in multiple environments.
\section{Discussion}
Here, a simple robot and task were employed to investigate the entire loss landscape of the weight manifold at high resolution. In our sweep across policy parameter sets for many robot designs, the results suggested that changes in one element of a robot's design (sensor distribution) greatly altered the loss surface. We hypothesized that this is likely to influence sample efficiency during policy optimization for a given design, and that design's resistance to catastrophic interference. This hypothesis was strengthened by the observation that changing sensor location induced changes in the number and placement across the loss surface of specialized parameter sets optimal for one environment, as well as the size of specialist overlap regions containing optimal generalized parameter sets that confer resistance to catastrophic interference.

We confirmed this hypothesis for four orthogonal training methods: we found that a robot design's learnability and resistance scores correlated with its sample efficiency, for this phototaxis task.

Lastly, we demonstrated that these high scoring designs could be found via simultaneous co-optimization of robot design and control, and that doing so was significantly more sample efficient than optimizing just the policy for a manually designed baseline robot. We concluded with evidence showing sensor homeostasis increased as robot designs improved, even though the loss function does not reward homeostasis. This suggests a causal link between agent design and sample efficient policy search: some designs better enable a robot to move such that observations converge across seemingly different environments, and allow it to perform the same, appropriate actions when aligned appropriately in each environment.

Even in the seemingly trivial case of phototaxis with contralateral connections on a minimal robot as first described by \citet{braitenberg1986vehicles}, we found that when challenged to perform in multiple environments, adjustments in design, specifically to sensor location, could either suppress or exacerbate catastrophic interference by expanding or shrinking the overlap of optimal policy parameters for that design across different task environments.

The physical location of sensors is thus a relevant property of robots nevertheless abstracted away in the mostly disembodied systems that address catastrophic interference reported in the literature to date. While sensor location could in principle be dynamically controlled via a lattice of sensors \cite{kramer2011wearable} or adjustable antenna \cite{fend2003active}, change in (and rational control over) other design attributes such as geometry \cite{kriegman2019automated}, material properties \cite{narang2018transforming}, or the number and placement of actuators \cite{lipson2000automatic} is much more difficult in practice, and such design elements are almost always presupposed and fixed prior to training \cite{cheney2018scalable}.

Overall we found that machines with the right design may use a single policy to succeed at a set of tasks that appear different to a robot with an ill suited body plan. The degree to which design influences learnability in more complex robots, task environments and behaviors has yet to be investigated, but will be the focus of future work.
\section{Limitations and Societal Impacts}
The main limitation of this work is the use of evolutionary methods for robot design discovery. The primary drawback of using evolution methods in this manner is the potential for perverse instantiation: there is a broader range of movement strategies for robots with arbitrary physical architecture, leading to increased surprise among loss minimizing solutions \citep{lehman2018surprising}. This is in general a problem for evolutionary algorithms but becomes compounded when aspects of physical design are added to training. If future extensions of this work include building physical instances of simulated robot designs, and those physical machines are employed to work in close proximity to humans, the breadth of surprising movements brings a risk of physical harm or death \cite{aliman2019augmented}.

Another limitation of evolutionary algorithms are their significant computational costs, which in turn can lead to a significant carbon footprint. However, here we show that finding good robot designs can improve policy discovery efficiency and reduce the amount of compute, suggesting that future versions of this method may be capable of reducing total compute requirements. Regardless, advances in differential physical simulation \cite{hu2019difftaichi} suggest that it may be possible to apply and follow gradients in the space of all possible robot designs, rendering both agent design and policy optimization compute and carbon efficient.

\begin{ack}
This work was supported by NSF EFRI award 1830870. Computational Resources were provided by The Vermont Advanced Computing Core (VACC). Thanks is also due to the beautiful wife of the first author whose continual support makes work like this possible.
\end{ack}

\bibliographystyle{plainnat}
\bibliography{main.bib}

\section*{Checklist}

%%% BEGIN INSTRUCTIONS %%%
The checklist follows the references.  Please
read the checklist guidelines carefully for information on how to answer these
questions.  For each question, change the default \answerTODO{} to \answerYes{},
\answerNo{}, or \answerNA{}.  You are strongly encouraged to include a {\bf
justification to your answer}, either by referencing the appropriate section of
your paper or providing a brief inline description.  For example:
\begin{itemize}
  \item Did you include the license to the code and datasets? \answerYes{See Section~\ref{gen_inst}.}
  \item Did you include the license to the code and datasets? \answerNo{The code and the data are proprietary.}
  \item Did you include the license to the code and datasets? \answerNA{}
\end{itemize}
Please do not modify the questions and only use the provided macros for your
answers.  Note that the Checklist section does not count towards the page
limit.  In your paper, please delete this instructions block and only keep the
Checklist section heading above along with the questions/answers below.
%%% END INSTRUCTIONS %%%

\begin{enumerate}

\item For all authors...
\begin{enumerate}
  \item Do the main claims made in the abstract and introduction accurately reflect the paper's contributions and scope?
    \answerYes{}
  \item Did you describe the limitations of your work?
    \answerYes{}
  \item Did you discuss any potential negative societal impacts of your work?
    \answerYes{}
  \item Have you read the ethics review guidelines and ensured that your paper conforms to them?
    \answerYes{}
\end{enumerate}

\item If you are including theoretical results...
\begin{enumerate}
  \item Did you state the full set of assumptions of all theoretical results?
    \answerNA{}
	\item Did you include complete proofs of all theoretical results?
    \answerNA{}
\end{enumerate}

\item If you ran experiments...
\begin{enumerate}
  \item Did you include the code, data, and instructions needed to reproduce the main experimental results (either in the supplemental material or as a URL)?
    \answerYes{The code is available on github and the url will be included if accepted for publication, it is being withheld here for anonymous review}
  \item Did you specify all the training details (e.g., data splits, hyperparameters, how they were chosen)?
    \answerYes{see \S\ref{sec:training}}
	\item Did you report error bars (e.g., with respect to the random seed after running experiments multiple times)?
    \answerYes{}
	\item Did you include the total amount of compute and the type of resources used (e.g., type of GPUs, internal cluster, or cloud provider)?
    \answerYes{see \S\ref{sec:overview}}
\end{enumerate}

\item If you are using existing assets (e.g., code, data, models) or curating/releasing new assets...
\begin{enumerate}
  \item If your work uses existing assets, did you cite the creators?
    \answerNA{}
  \item Did you mention the license of the assets?
    \answerNA{}
  \item Did you include any new assets either in the supplemental material or as a URL?
    \answerNA{}
  \item Did you discuss whether and how consent was obtained from people whose data you're using/curating?
    \answerNA{}
  \item Did you discuss whether the data you are using/curating contains personally identifiable information or offensive content?
    \answerNA{}
\end{enumerate}

\item If you used crowdsourcing or conducted research with human subjects...
\begin{enumerate}
  \item Did you include the full text of instructions given to participants and screenshots, if applicable?
    \answerNA{}
  \item Did you describe any potential participant risks, with links to Institutional Review Board (IRB) approvals, if applicable?
    \answerNA{}
  \item Did you include the estimated hourly wage paid to participants and the total amount spent on participant compensation?
    \answerNA{}
\end{enumerate}

\end{enumerate}

\end{document}